\documentclass[10pt,twocolumn,letterpaper]{article}

\usepackage{iccv}
\usepackage{times}
\usepackage{epsfig}
\usepackage{graphicx}
\usepackage{amsmath}
\usepackage{amssymb}
\usepackage{bbm}
\usepackage{mathtools}
\usepackage{booktabs}
\usepackage{textcomp}
\usepackage{adjustbox}
\usepackage{multirow}
\usepackage{hhline}
\usepackage{siunitx}
\usepackage{enumitem}
\newcommand{\tabincell}[2]{\begin{tabular}{@{}#1@{}}#2\end{tabular}}

\usepackage[accsupp]{axessibility}  % Improves PDF readability for those with disabilities.

\usepackage{mathtools}

\DeclarePairedDelimiter\floor{\lfloor}{\rfloor}

\usepackage{color, colortbl}
\definecolor{LightCyan}{rgb}{0.88,1,1}
\definecolor{LightRed}{rgb}{1,0.88,0.95}
\definecolor{LightGreen}{rgb}{0.56, 0.93, 0.56}
\definecolor{gray}{rgb}{0.95, 0.95, 0.95}

\usepackage{amssymb}% http://ctan.org/pkg/amssymb
\usepackage{pifont}% http://ctan.org/pkg/pifont

\renewcommand{\baselinestretch}{0.98}

\usepackage[pagebackref=true,breaklinks=true,letterpaper=true,colorlinks,bookmarks=false]{hyperref}

\iccvfinalcopy % *** Uncomment this line for the final submission

\ificcvfinal\fi

\begin{document}

%%%%%%%%% TITLE
\title{Guided Point Contrastive Learning for \\ Semi-supervised Point Cloud Semantic Segmentation}

\author{
	Li Jiang$^{1}$ \quad Shaoshuai Shi$^{1}$ \quad Zhuotao Tian$^{1}$ \quad Xin Lai$^{1}$ \quad Shu Liu$^{2}$ \quad Chi-Wing Fu$^{1}$ \quad Jiaya Jia$^{1,2}$\\
	$^{1}$The Chinese University of Hong Kong \quad $^{2}$SmartMore\\
	{\tt\small \{lijiang, zttian, xinlai, cwfu, leojia\}@cse.cuhk.edu.hk} \\
	\tt\small  ssshi@ee.cuhk.edu.hk \quad \tt\small sliu@smartmore.com
}

\renewcommand{\baselinestretch}{0.98}

\maketitle
% Remove page # from the first page of camera-ready.
\ificcvfinal\thispagestyle{empty}\fi

%%%%%%%%% ABSTRACT
\begin{abstract}
	Rapid progress in 3D semantic segmentation is inseparable from the advances of deep network models, which highly rely on large-scale annotated data for training. To address the high cost and challenges of 3D point-level labeling, we present a method for semi-supervised point cloud semantic segmentation to adopt unlabeled point clouds in training to boost the model performance. Inspired by the recent contrastive loss in self-supervised tasks, we propose the guided point contrastive loss to enhance the feature representation and model generalization ability in semi-supervised setting. Semantic predictions on unlabeled point clouds serve as pseudo-label guidance in our loss to avoid negative pairs in the same category. Also, we design the confidence guidance to ensure high-quality feature learning. Besides, a category-balanced sampling strategy is proposed to collect positive and negative samples to mitigate the class imbalance problem. Extensive experiments on three datasets (ScanNet V2, S3DIS, and SemanticKITTI) show the effectiveness of our semi-supervised method to improve the prediction quality with unlabeled data. 
\end{abstract}

%%%%%%%%% BODY TEXT

%------------------------------------------------------------------------
%%%%%%% introduction
\vspace{-2mm}
\section{Introduction}

3D point cloud semantic segmentation is a fundamental and essential perception task for many downstream applications~\cite{lahoud20193d, shi2019pointrcnn,yan2020sparse, jiang2020pointgroup}. 
Existing deep-learning-based methods for the task heavily rely on the availability and quantity of labeled point cloud data for the model training.
However, 3D point-level labeling is time-consuming and labor-intensive.
Compared with point cloud labeling, point cloud collection requires much less effort, mainly by means of 3D scanning followed by some data post-processing.
Hence, we are motivated to explore semi-supervised learning (SSL) for improving the data efficiency and performance of deep segmentation models with unlabeled point clouds.

\begin{figure}
	\centering
	\scalebox{0.95}[0.86]{
		\begin{tabular}{@{\hspace{0.0mm}}c@{\hspace{3.0mm}}c@{\hspace{3.0mm}}c}
			\includegraphics[width=0.3\linewidth]{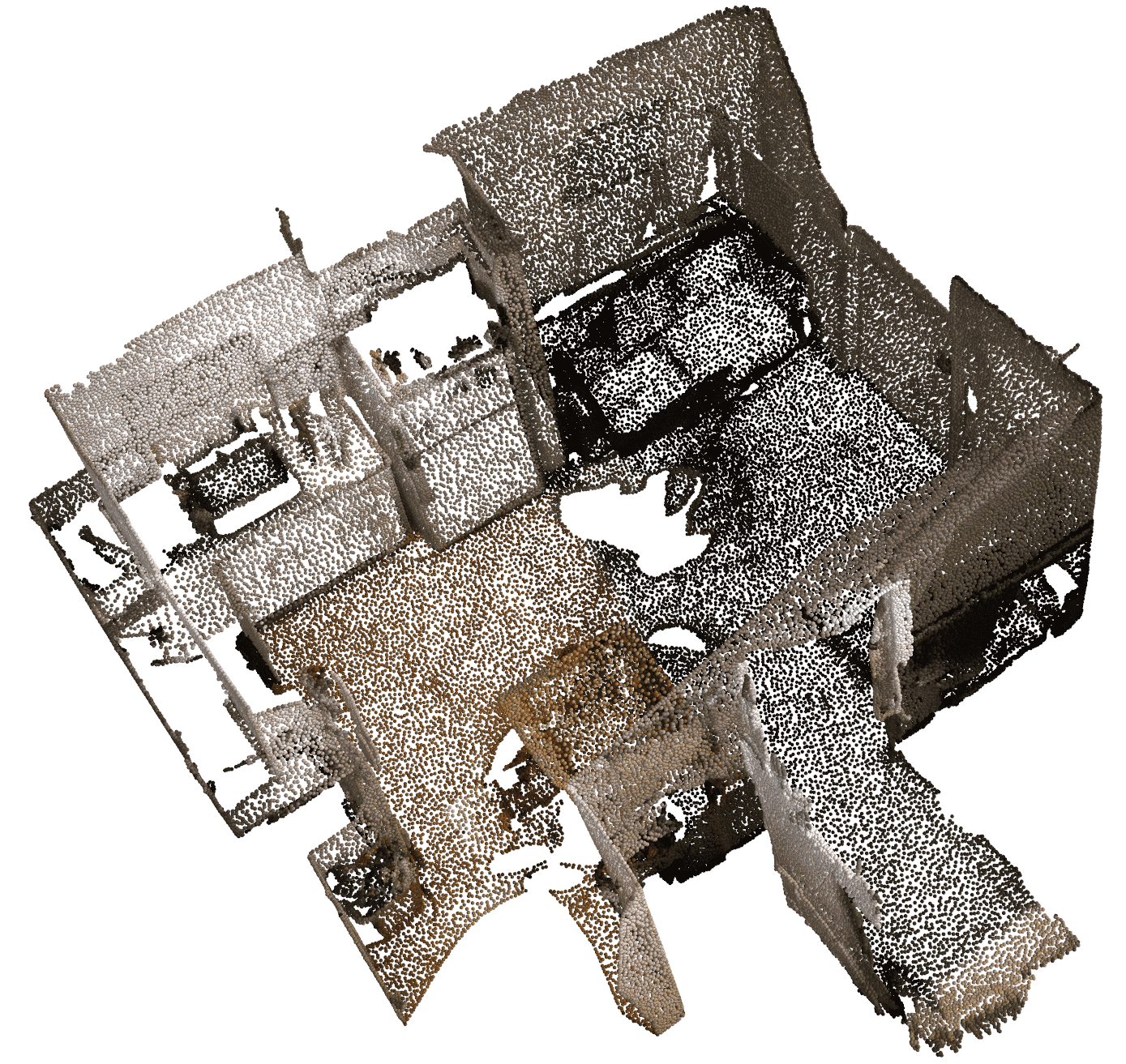}& 
			\includegraphics[width=0.28\linewidth]{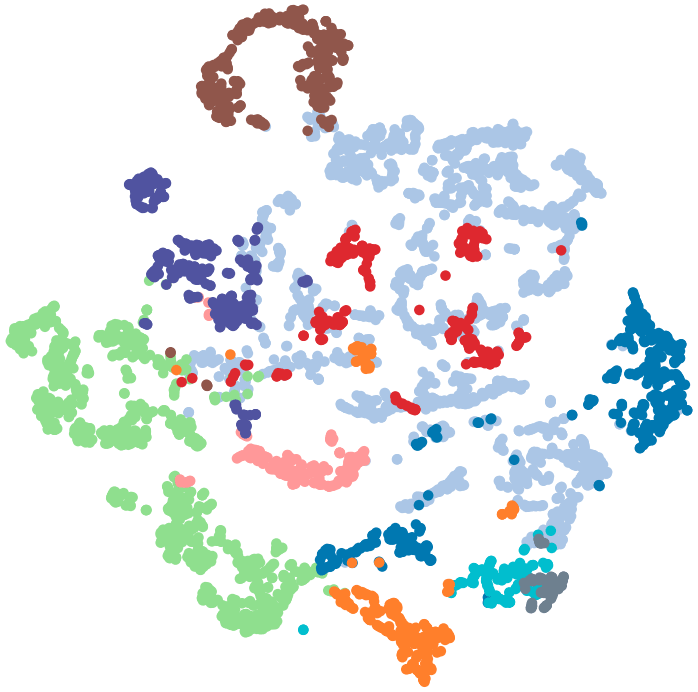} & 
			\includegraphics[width=0.28\linewidth]{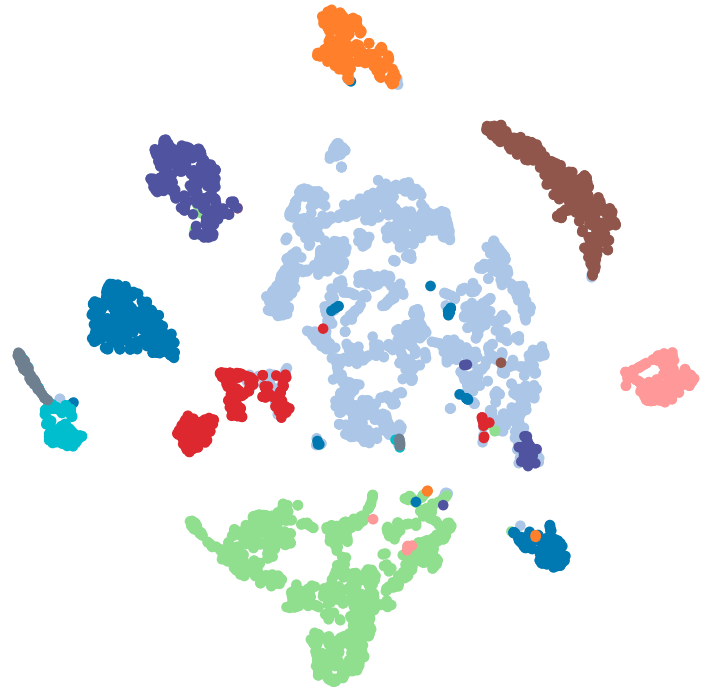} \\
			\includegraphics[width=0.3\linewidth]{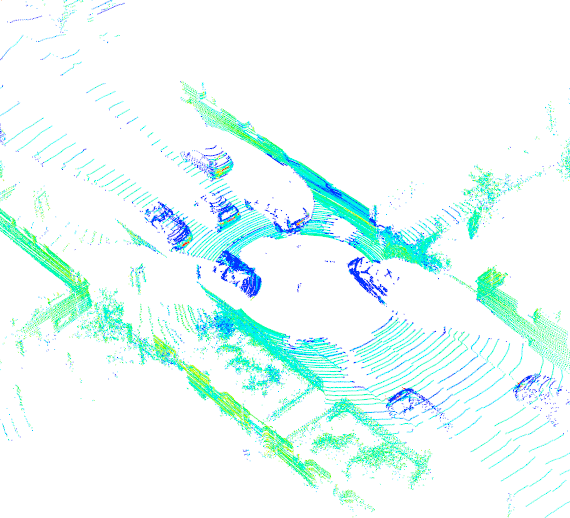} & 
			\includegraphics[width=0.28\linewidth]{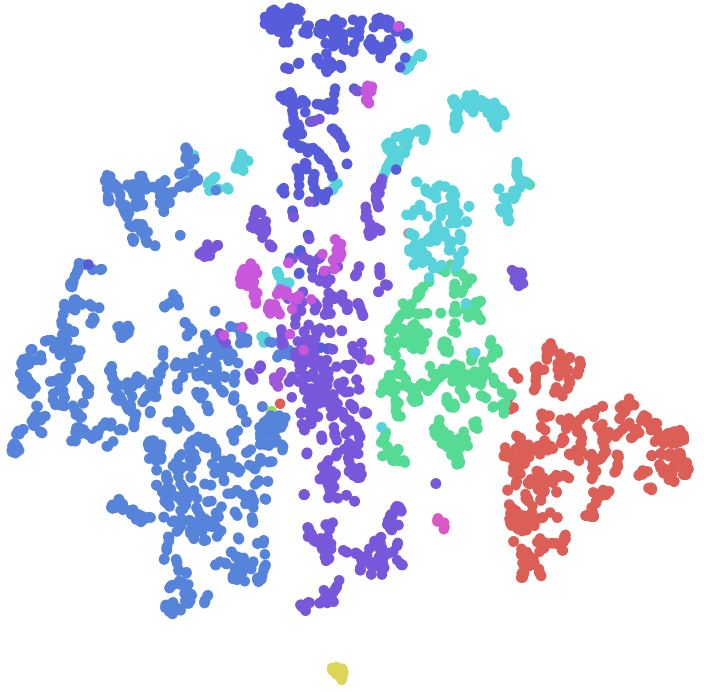} & 
			\includegraphics[width=0.28\linewidth]{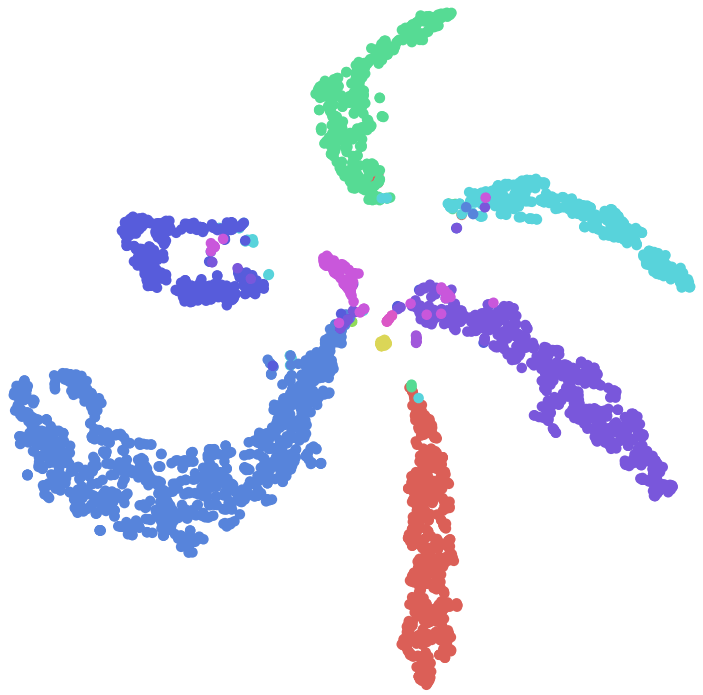} \\
			{\scriptsize Input Point Cloud} & {\scriptsize Without Pseudo Guidance} & {\scriptsize With Our Pseudo Guidance}\\
	\end{tabular}}
	\caption{Visualizations of feature embeddings for unlabeled data under different loss strategies, separately trained on the indoor ScanNet V2~\cite{dai2017scannet} (top) and outdoor SemanticKITTI~\cite{behley2019semantickitti} (bottom) with 20\% labeled data. 
		Our proposed pseudo guidance enhances the feature learning, with feature embeddings in different categories being better separated (middle vs. right columns).
	}
	\label{fig:teaser}
	\vspace{-5mm}
\end{figure}

While SSL has been widely explored for tasks on 2D images~\cite{NIPS2004_96f2b50b,temporalICLR2017,NIPS2017_68053af2,miyato2018virtual,zhai2019s4l,ouali2020semi,ke2020guided}, it is rather underexplored for 3D point clouds.
To achieve SSL, a common strategy is consistency regularization~\cite{temporalICLR2017,NIPS2017_68053af2}, which aligns features of the same image/pixel under different perturbations for maintaining the prediction consistency when exploiting unlabeled data.
Our method shares this common ground in SSL by encouraging similar and robust features for matched 3D point pairs with different transformations.
Yet, inspired by the contrastive loss applied in self-supervised learning~\cite{hadsell2006dimensionality,ye2019unsupervised,he2020momentum,chen2020simple}, we further enhance the feature representation by proposing the {\em guided point contrastive loss\/} to additionally enlarge the distance between inter-category features by using the semantic predictions as guidance in the semi-supervised setting.

Contrastive learning starts with works on 2D images, and is recently extended by PointContrast~\cite{xie2020pointcontrast} to 3D point clouds as a pre-training pretext task in a  self-supervised setting.
The point contrastive loss encourages the matched positive point pairs to be similar in the embedding space while pushing away the negative pairs. Yet, without any label, negative pairs in the same category may also be sampled, especially for large objects (e.g., sofa) and redundant background classes (e.g., floor and wall); these negative pairs actually weaken the features' discriminative ability. 
Unlike PointContrast, we leverage a few labeled point clouds to optimize the network model for producing point-level semantic predictions, and meanwhile, utilize the predicted semantic scores and labels for the unlabeled data to guide the contrastive loss computation. Our {\em pseudo-label guidance} helps alleviate the side effect of intra-class negative pairs in feature learning, while our {\em confidence guidance} utilizes the semantic scores to reduce the chance of feature worsening. 
Also, we propose a {\em category-balanced sampling} strategy to exploit pseudo labels to mitigate the class imbalance issue in point sampling, helping to preserve point samples from rare categories and to improve the feature diversity in contrastive learning.
As revealed in the t-SNE visualizations in Fig.~\ref{fig:teaser}, the model equipped with our pseudo guidance learns more discriminative point-wise features.

We follow the conventional practice in SSL to conduct experiments with a small portion of labeled data and  a larger portion of unlabeled data and then evaluate how effective an SSL method improves the performance with the unlabeled data.
Excellent performance for both indoor (ScanNet V2~\cite{dai2017scannet} and S3DIS~\cite{armeni20163d}) and outdoor (SemanticKITTI~\cite{behley2019semantickitti}) scenes are obtained, showing the effectiveness of our semi-supervised method, which surpasses the supervised-only models with 5\%, 10\%, 20\%, 30\%, and 40\% labeled data by a large margin consistently on all three datasets.
Also, we experiment with 100\% labeled data, in which the labeled set is also fed into the unsupervised branch with our guided point contrastive loss as an auxiliary feature learning loss. 
In this case, the accuracy of our method still exceeds the baseline with only supervised cross entropy loss, showing that without extra unlabeled data, our guided point contrastive loss also helps to refine the feature representation and model's discriminative ability.

Our contributions are threefold:
\begin{itemize}
	\vspace*{-1.25mm}
	\item
	We adopt semi-supervised learning to 3D scene semantic segmentation, demonstrating that unlabeled point clouds can help to enhance the feature learning in both indoor and outdoor scenes.
	\vspace*{-1.25mm}
	\item
	We extend contrastive learning to 3D point cloud semi-supervised semantic segmentation with pseudo-label guidance and confidence guidance. 
	\vspace*{-1.25mm}
	\item
	We propose a category-balanced sampling strategy to alleviate the point class imbalance issue and to increase the embedding diversity.
\end{itemize}

%%%%%%% RW
\section{Related Works}

%%%%%%%%%%%%%%%%%%%%%%%%%%%%%%%%%%%%%%%%%

\noindent
{\bf Point cloud segmentation.} \
Various approaches have been explored for 3D semantic segmentation.
Voxel-based approaches~\cite{maturana2015voxnet, song2017semantic} utilize 3D convolutional neural networks by transforming irregular point clouds to regular 3D grids.
Other approaches explore the sparsity of voxels for high-resolution 3D representations with OctNet~\cite{riegler2017octnet} or sparse convolution~\cite{3DSemanticSegmentationWithSubmanifoldSparseConvNet,choy20194d}.
Pioneered by PointNet~\cite{qi2017pointnet,NIPS2017_d8bf84be}, point-based approaches directly learn point features from raw point clouds with assorted hierarchical local feature aggregation strategies~\cite{zhao2019pointweb,wu2019pointconv,jiang2019hierarchical}.
KPConv~\cite{thomas2019kpconv} defines a kernel function on points for conducting convolutions on local points.
There are also works, e.g.,~\cite{simonovsky2017dynamic,landrieu2018large,wang2019dynamic}, that incorporate graph convolutions for point feature learning.

To train the network model, these fully-supervised approaches require data with point-wise labels, which are time-consuming and tedious to prepare as well as error-prone.
Hence, in this work, we incorporate unlabeled point clouds in network training for improving the data efficiency in 3D point cloud semantic segmentation.

%%%%%%%%%%%%%%%%%%%%%%%%%%%%%%%%%%%%%%%%%

\vspace*{1mm}
\noindent
{\bf Semi-supervised learning}
(SSL) aims to improve a model by learning from unlabeled data, in addition to labeled data.
Existing works on SSL mainly focus on image classification~\cite{NIPS2004_96f2b50b, lee2013pseudo, temporalICLR2017, NIPS2017_68053af2, miyato2018virtual, zhai2019s4l} and image semantic segmentation~\cite{souly2017semi, hung2018adversarial, mittal2019semi, ouali2020semi, ke2020guided, french2020semi, zou2021pseudoseg}.
Consistency regularization is a common strategy for SSL, emphasizing that the model predictions should be consistent for different perturbations applied to the same input.
$\Pi$-model~\cite{temporalICLR2017}, a simplified version of $\Gamma$-model~\cite{NIPS2015_378a063b}, encourages consistent model outputs for different dropouts and augmentations on the same input, while temporal ensembling~\cite{temporalICLR2017} and Mean Teacher~\cite{NIPS2017_68053af2} adopt the exponential moving average strategy to stabilize the predictions for consistency regularization.

Recent SSL methods for image segmentation show that pixel-wise consistency could be achieved by perturbing the input images~\cite{french2020semi} or the intermediate features~\cite{ouali2020semi}, or by feeding the same image to different models~\cite{ke2020guided}.
Pseudo-label-based self-training~\cite{lee2013pseudo,zou2021pseudoseg} is another approach for SSL, in which we first train a model with labeled data then refine it by generating pseudo labels on the unlabeled data for further training. 
Some other works~\cite{souly2017semi, hung2018adversarial} also adopt a generative adversarial network for SSL image segmentation to incorporate unlabeled images for learning. 

Though many SSL works have been proposed for images, SSL for point cloud scenes is rather underexplored.
Currently, there are two works for 3D detection that leverage unlabeled scenes by Mean-Teacher framework~\cite{zhao2020sess} or by quality-aware pseudo labeling~\cite{wang20203dioumatch}.
Compared with 3D box annotations, point-wise dense annotations for 3D point cloud segmentation are more resource-intensive. Hence, we propose a novel SSL framework for the task, demonstrating the feasibility of incorporating unlabeled point clouds to improve the performance of segmenting 3D points.

%%%%%%%%%%%%%%%%%%%%%%%%%%%%%%%%%%%%%%%%%

\vspace*{1mm}
\noindent
{\bf Contrastive Learning} is a widely-used approach for unsupervised learning~\cite{hjelm2018learning,ye2019unsupervised,he2020momentum,chen2020simple,NEURIPS2020_fcbc95cc}.
Its core idea is the contrastive loss~\cite{hadsell2006dimensionality} that encourages the features of the query samples to be similar to those of the positive key samples, while being dissimilar with those of the negative key samples.
A common choice of contrastive loss is InfoNCE~\cite{oord2018representation}, which measures the similarity by a dot product.
PointContrast~\cite{xie2020pointcontrast} proposes the PointInfoNCE loss for point-level unsupervised representation learning,  and their follow-up work~\cite{hou2020exploring} proposes a ShapeContext-like spatial partition for location-aware contrastive learning.
Supervised contrastive learning~\cite{NEURIPS2020_d89a66c7} is also proposed recently to better align the intra-class features with labeled data.
In this work, we extend contrastive learning to support semi-supervised point cloud segmentation and propose to incorporate point-wise pseudo labels to the contrastive loss for better distinguishing the positive and negative samples and collectively using both the labeled and unlabeled point clouds for learning a more effective representation.

%%%%%%%%% method

\begin{figure*}
	\centering
	\includegraphics[width=0.95\linewidth, height=6.5cm]{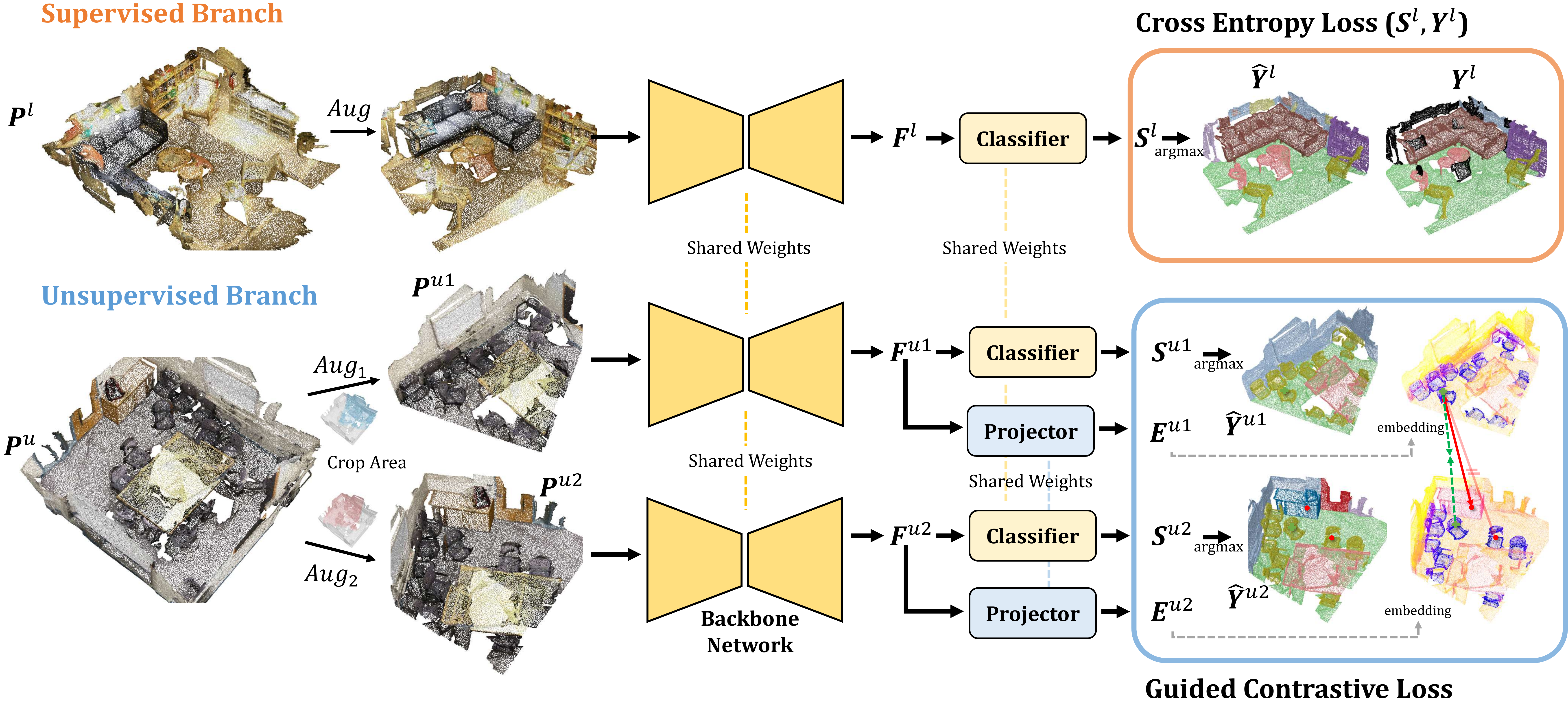}
	\vspace{-1mm}
	\caption{Our SSL framework. The superscripts $l$ and $u$ stand for `labeled' and `unlabeled', respectively.  $\mathbf{P}$ is the input point cloud. 	$Aug$ represents `augmentation' (\eg, crop). $\mathbf{P}^u$ is independently augmented to form $\mathbf{P}^{u1}$ and $\mathbf{P}^{u2}$, as input to the unsupervised branch. $\mathbf{F}$ is the output point-wise features of the backbone U-Net, which is further fed to a classifier to predict semantic scores $\mathbf{S}$.  In the unsupervised branch, $\mathbf{F}$ is also fed to a projector to produce feature embedding $\mathbf{E}$. $\hat{\mathbf{Y}}$ stands for the point-wise class predictions, whereas $\mathbf{Y}$ is the ground-truth labels. The weights of the backbone network, classifier, and projector are shared for all input point clouds. Cross entropy loss constrains the supervised training with $\mathbf{Y}^l$, whereas our guided contrastive loss steers the feature learning in the unsupervised branch.}
	\vspace{-4mm}
	\label{fig:ssl_framework}
\end{figure*}

\section{Our Approach}

\subsection{Preliminaries on Point-level Feature Learning}

The key in SSL is to learn feature representation from unlabeled data, which is a common goal shared by both unsupervised learning~\cite{he2020momentum,chen2020simple} and SSL~\cite{NIPS2017_68053af2,temporalICLR2017,ouali2020semi}. When it comes to 3D semantic segmentation, feature learning in point level is of critical importance. Hence, we begin this section by revisiting and analyzing label-free point-level feature learning in an unsupervised setting and then in SSL.

%%%%%%%%%%%%%%%%%%%%%%%%%%%%%%%%%%%%%%%%%%%

\vspace*{1mm}
\noindent
{\bf Point-level contrast in self-supervised learning.} 
PointContrast~\cite{xie2020pointcontrast} firstly proposes the point-level self-supervised strategy for pre-training with unlabeled point clouds. 
It extends the InfoNCE loss~\cite{oord2018representation} to points as PointInfoNCE loss for contrastive learning on 3D scenes:
\vspace{-2.1mm}
\begin{equation}
\small
L_{pc} = -\dfrac{1}{| \mathbf{M}_p|}\sum_{(i, j) \in \mathbf{M}_p}\log\dfrac{\exp(\mathbf{E}^{u1}_i \cdot \mathbf{E}^{u2}_j / \tau)}{\sum_{(\cdot, k) \in \mathbf{M}_p} \exp(\mathbf{E}^{u1}_i \cdot \mathbf{E}^{u2}_k / \tau)},
\end{equation}
where $\mathbf{M}_p$ is the index set of randomly-sampled positive pairs (one-to-one matched points) across two point clouds perturbed from the same input; $\mathbf{E}^{u1}$ and $\mathbf{E}^{u2}$ are feature embeddings of the two point clouds; and $\tau$ is a temperature hyperparameter.
For point $i$ in the first point cloud $u1$, $(i, j) \in \mathbf{M}_p$ is a positive pair, whose feature embeddings $(\mathbf{E}^{u1}_i, \mathbf{E}^{u2}_j)$ are encouraged to be similar, while $\{(i, k) | (\cdot, k) \in \mathbf{M}_p, k$$\neq$$j\}$ are negative point 
pairs.  Point $i$ is called the anchor point; its feature embedding is enforced to be dissimilar with the feature embeddings of all its negative points.
PointContrast serves as a pretext task for pre-training and validates the effectiveness of point-level contrastive loss in point cloud self-supervised learning.

%%%%%%%%%%%%%%%%%%%%%%%%%%%%%%%%%%%%%%%%%%%

\vspace*{1mm}
\noindent
{\bf Point-level consistency in SSL.}~
Consistency regularization is a widely-used strategy to exploit unlabeled data for enhancing feature robustness. 
Hence, we define a simple baseline with consistency regularization. 
For point-level consistency, inspired by 2D SSL~\cite{NIPS2017_68053af2}, one may enforce a corresponding point pair with different augmentations to have similar feature representation by minimizing the Mean-Squared Error (MSE) between the feature embeddings of the points.
Formally, the loss in the unsupervised branch in SSL with MSE can be expressed as
\vspace{-1.5mm}
\begin{equation}
\label{eq:mse}
L_u = \dfrac{1}{|\mathbf{M}|}\sum_{(i, j) \in \mathbf{M}}{\left\| \mathbf{E}^{u1}_i - \mathbf{E}^{u2}_j\right\|^2},
\end{equation}
\vspace{-3mm}

\noindent
where $\mathbf{M}$ is the index set of all matched point pairs across point clouds $u1$ and $u2$ perturbed from the same input. In SSL, we combine $L_u$ with the following supervised cross entropy loss $L_l$ on labeled data for model training:
\vspace{-3mm}
\begin{equation}
\label{eq:ce}
L_l = \dfrac{1}{N^l}\sum_{i = 1}^{N^l} (-{\mathbf{S}_{i}^l[\mathbf{Y}_i^l]} + \log{\sum\nolimits_j \exp \mathbf{S}_i^l[j]}),
\end{equation}
\vspace{-3mm}

\noindent where $N^l$ is the number of points in the given labeled point cloud; $\mathbf{S}^l$ is the predicted semantic scores; and $\mathbf{Y}^l$ represents the ground-truth labels.

%%%%%%%%%%%%%%%%%%%%%%%%%%%%%%%%%%%%%%%%%%%

\vspace*{1mm}
\noindent
{\bf Discussion.}~
Though both PointInfoNCE and our SSL baseline could learn from unlabeled point clouds and benefit 3D semantic segmentation (see Table~\ref{tab:exp_ab3}), they have several drawbacks:
(i) \textit{Negative point pairs of same category could worsen the feature learning}:
In the unsupervised setting of PointInfoNCE, a negative point pair $(i, k)$ may come from the same semantic category, so pushing away their embeddings $(\mathbf{E}^{u1}_i, \mathbf{E}^{u2}_k)$ could degrade the feature learning. 
(ii) \textit{Points from the same category are likely to be sampled, especially for large objects or for common categories such as road}:
Random sampling could easily produce unfavorable negative point pairs that actually come from the same category.
(iii) \textit{Feature distance for both intra- and inter-class should be considered}:
In our SSL baseline, only paired intra-class features are constrained to be similar. 
However, the inter-class feature distance should also be enlarged to better improve the semantic segmentation.

To mitigate the above problems, we focus on exploring and leveraging the information from labeled point clouds to better guide the feature learning from unlabeled point clouds for improving the 3D scene semantic segmentation.

%%%%%%%%%%%%%%%%%%%%%%%%%%%%%%%%%%%%%%%%%%%
\vspace{-1mm}
\subsection{Pseudo Guidance on Contrastive Learning}
\vspace{-1.5mm}

Now, we focus on the setting of semi-supervised learning (SSL) for 3D point cloud semantic segmentation, in which we could  leverage some labeled data to train the model to produce semantic predictions for unlabeled scenes.
We accordingly propose the {\em Guided Point Contrastive Learning framework\/} for SSL-based point cloud segmentation and leverage the semantic predictions as pseudo guidance for improving the contrastive learning on unlabeled point clouds.
Fig.~\ref{fig:ssl_framework} shows the overall architecture of our framework, which consists of a supervised branch and an unsupervised branch. In this section, we focus on our guided contrastive loss in the unsupervised branch.

Formally, for a pair of perturbed point clouds $(\mathbf{P}^{u1}, \mathbf{P}^{u2})$ from the same unlabeled data, we can generate their pseudo labels $(\hat{\mathbf{Y}}^{u1}, \hat{\mathbf{Y}}^{u2})$ and label confidences $(\mathbf{C}^{u1}, \mathbf{C}^{u2})$ from the predicted semantic scores $(\mathbf{S}^{u1}, \mathbf{S}^{u2})$ as follows:
\vspace{-2mm}
\begin{equation} \label{eq:score_max}
\hat{\mathbf{Y}}^{*}_i = \arg\max ~\mathbf{S}^{*}_i,~~~~~  \mathbf{C}^{*}_i = \max~\sigma(\mathbf{S}^{*}_i),
\end{equation}
\vspace{-5mm}

\noindent
where $*$ is $u1$ or $u2$ and $\sigma$ denotes the softmax function. 

We then denote $\mathbf{M}_p$ as the set of matched positive point pairs across point clouds $u1$ and $u2$ perturbed from the same input. 
For the negative point sets, instead of using points in $\mathbf{M}_p$ as in~\cite{xie2020pointcontrast}, we separately sample negative points to ensure negative samples 
can also come
from un-matched regions. 
We denote the negative point sets sampled from point clouds $u1$ and $u2$ as $\mathbf{M}_n^{u1} \subseteq \{1, 2, ..., N^{u1}\}$ and $\mathbf{M}_n^{u2} \subseteq \{1, 2, ..., N^{u2}\}$, respectively, 
where $N^{u1}$ and $N^{u2}$ denote the number of points in the associated point clouds.

\vspace{1mm}
\noindent 
{\bf Guided contrastive loss. }~
Given the positive point pair set $\mathbf{M}_p$ and negative point sets $\mathbf{M}_n^{u1}$ and $\mathbf{M}_n^{u2}$, our guided contrastive loss $L_u^{(i, j)}$ for the positive point pair $(i, j) \in \mathbf{M}_p$ could be represented as 
\vspace{-2mm}
\begin{equation}\label{eq:main_loss}
\small
\begin{cases}
L_{u1}^{(i, j)} =
-\log
\dfrac{\exp(\mathbf{E}^{u1}_i \cdot~\text{stopgrad}\left(\mathbf{E}^{u2}_j\right)~/~ \tau)}{\sum\limits_{\substack{k \in \mathbf{M}_n^{u2} \cup \{j\}}} {G_{i,k}} \cdot \exp(\mathbf{E}^{u1}_i \cdot ~\text{stopgrad}(\mathbf{E}^{u2}_k) / \tau)},
\\ 
L_{u2}^{(i, j)} =
-\log
\dfrac{\exp(\mathbf{E}^{u2}_j \cdot~\text{stopgrad}\left(\mathbf{E}^{u1}_i\right)~/~ \tau)}{\sum\limits_{\substack{k \in \mathbf{M}_n^{u1} \cup \{i\}}} {G_{k,j}} \cdot \exp(\mathbf{E}^{u2}_j \cdot ~\text{stopgrad}(\mathbf{E}^{u1}_k) / \tau)},
\\
L_u^{(i, j)} = \mathbbm{1}{(\mathbf{C}^{u2}_{j} \ge \gamma)}\cdot L_{u1}^{(i, j)} +  \mathbbm{1}{(\mathbf{C}^{u1}_{i} \ge \gamma)}\cdot L_{u2}^{(i, j)},
\end{cases}
\end{equation}
where $\gamma$ is a confidence threshold. 
Note that the loss is computed on $\mathbf{P}^{u1}$ and $\mathbf{P}^{u2}$ separately. 
For each side, the features from the other side is detached to stop gradients and is thus treated as constant references for better optimizing the features on the current side. 

\begin{figure}
	\centering
	\includegraphics[width=1.0\linewidth, height=5.8cm]{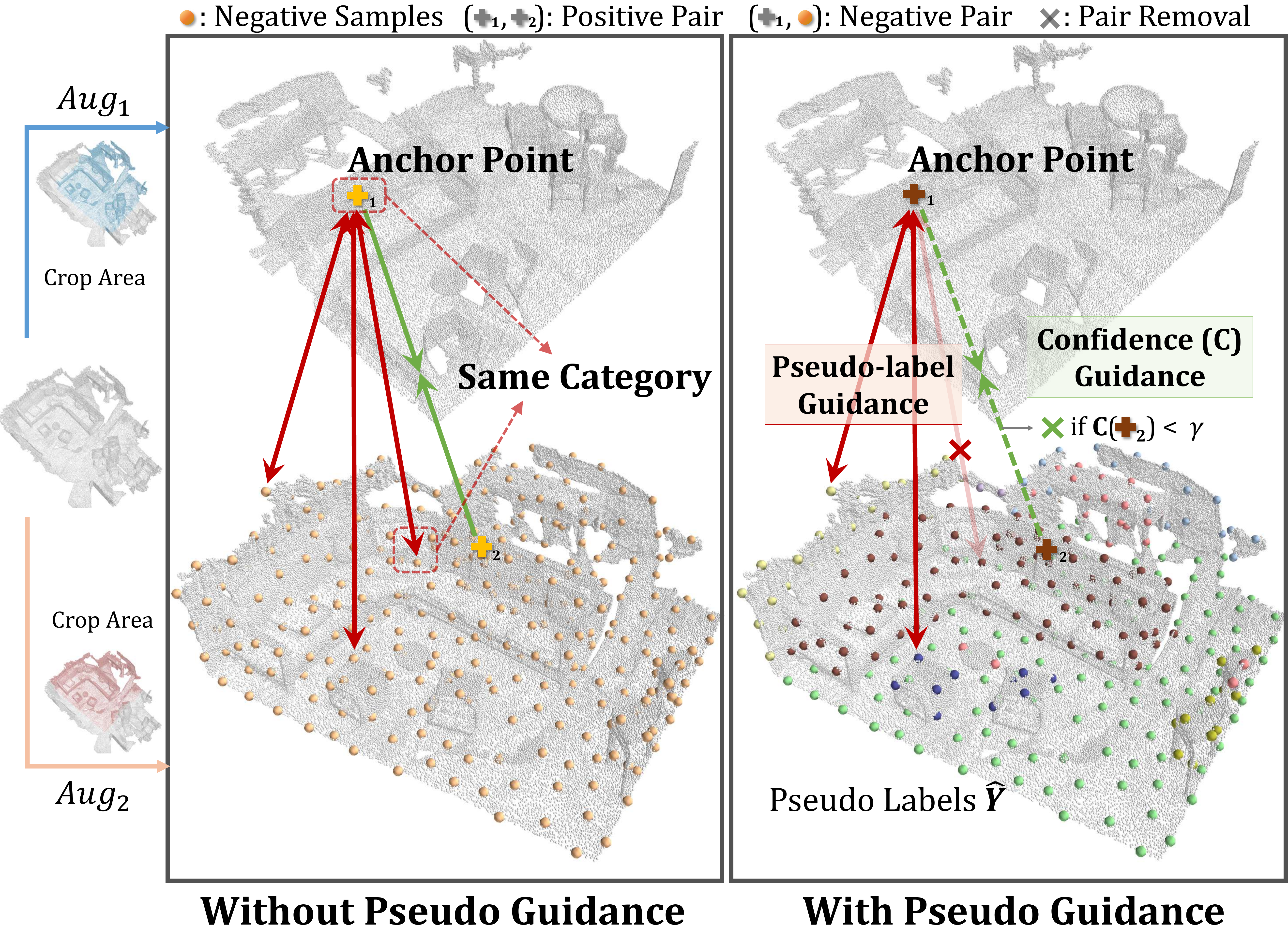}
	\caption{Pseudo Guidance on Contrastive Learning. 
		\textbf{Left}: The anchor point is encouraged to be similar to its matched positive point, while being dissimilar to all the negative samples. 
		\textbf{Right}: Semantic predictions (shown in different colors for different categories) are incorporated  into the contrastive loss design. {\em Pseudo-label guidance} filters out the negative samples that have same pseudo label as the anchor point.
		{\em Confidence guidance} adds a constraint on the positive point pair that the anchor point is avoided to be pulled close to a matched positive point with low confidence.}
	\vspace{-5mm}
	\label{fig:pseudo_guidance}
\end{figure}

By leveraging the semantic predictions in Eq.~\eqref{eq:score_max}, we propose two pseudo guidances in Eq.~\eqref{eq:main_loss} to guide the feature learning from unlabeled point clouds, which are illustrated in Fig.~\ref{fig:pseudo_guidance} and discussed below: 
\begin{itemize}[leftmargin=*]
	\vspace{-1mm}
	\item \textit{Pseudo-label guidance}: $G$ is the pseudo-label guidance for filtering negative point pairs with the same pseudo labels,	which is defined as
	\vspace{-3mm}
	\begin{equation}
	G_{i, j}=
	\begin{cases}
	\mathbbm{1}{(\hat{\mathbf{Y}}^{u1}_i \neq \hat{\mathbf{Y}}^{u2}_j)}&, ~~\text{if}~(i, j)~\notin \mathbf{M}_p, \\
	1&, ~~\text{otherwise}.
	\end{cases}
	\end{equation}
	\vspace{-3mm}
	
	\noindent
	As shown in Fig.~\ref{fig:pseudo_guidance}, for an anchor point on `sofa', many negative samples are also on `sofa'. 
	Pushing away such an intra-category negative point pair could adversely affect the feature learning (Fig.~\ref{fig:pseudo_guidance} left). 
	By incorporating our proposed pseudo-label guidance on contrastive learning, only negative feature pairs with different semantic predictions are forced to be dissimilar (Fig.~\ref{fig:pseudo_guidance} right). 
	
	\vspace{-1mm}
	\item \textit{Confidence guidance}:
	Since feature $\mathbf{E}^{u1}_i $ is pulled to $\mathbf{E}^{u2}_j$ in $L_{u1}^{(i, j)}$, we additionally incorporate a { confidence guidance} $ \mathbbm{1}{(\mathbf{C}^{u2}_{j} \ge \gamma)}$ on $L_{u1}^{(i, j)}$ to avoid $\mathbf{E}^{u1}_i $ learning from a low-confidence feature; so is the same for $L_{u2}^{(i, j)}$. Confidence guidance largely prevents feature worsening and improves the feature learning quality. 
\end{itemize}

The overall guided contrastive loss is the average of the losses for the positive pairs in $\mathbf{M}_p$:
\vspace{-3mm}
\begin{equation}
L_u =  \dfrac{1}{|\mathbf{M}_p|}\sum_{(i, j) \in \mathbf{M}_p} L_u^{(i, j)}.
\end{equation}

%%%%%%%%%%%%%%%%%%%%%%%%%%%%%%%%%%%%%%%%%%%%
\vspace{-2mm}
\subsection{Category-balanced Sampling}
\label{sec:cbs}
The computation cost of the guided contrastive loss is highly correlated with the positive / negative point numbers.
As there is a large number of points in each point cloud (\eg, around 100k - 1000k for an indoor scene and around 100k for an outdoor LiDAR frame), usually we could not take all the points as positive or negative samples. 
Hence, for data with imbalanced category distribution, some categories with a very small number of points may have little chance to be sampled with random sampling, while some large-portion categories are often sampled redundantly, thus affecting the feature diversity in contrastive learning. To mitigate this problem, we propose a simple but effective sampling strategy---Category-Balanced Sampling (CBS). 

\vspace{1.5mm}
\noindent 
\textbf{Category-balanced sampling for positive pairs.}~
We define the category set as $\mathcal{C}$, so the number of categories is $|\mathcal{C}|$. 
For a pair of perturbed point clouds, we denote the set of all matched point pairs as $\mathbf{M}$ and reorganize $\mathbf{M}$ by the predicted categories of the $1^{st}$ points in the pairs.  The number of matched point pairs in each category $c \in \mathcal{C}$ is denoted as $N_M^c$.
To sample a total number of $K_p$ positive point pairs from $\mathbf{M}$ to form $\mathbf{M}_p$ for our guided contrastive loss, our CBS strategy evenly samples positive pairs from each category. Specifically, for category $c$, the number of selected positive pairs $N_{M_p}^{c}$ is calculated as
\vspace{-3mm}
\begin{align}
N_{M_p}^{c} =\min\{\floor*{\frac{K_p}{|\mathcal{C}|}}, N_M^c\}.
\end{align}
\vspace{-3mm}

\noindent
Then, additional $K_p - \sum_c N_{M_p}^{c}$ pairs are sampled from all categories to ensure the total number of positive pairs is $K_p$.

\vspace{1.5mm}
\noindent 
\textbf{Category-balanced sampling for negative point set.}~
To enhance the sample diversity of negative points, we conduct CBS by collecting negative samples from scenes in the entire training set instead of just from the current scene, since some categories may even be absent in a specific scene or batch. 
Precisely, we maintain a category-aware negative embedding memory bank of size $|\mathcal{C}| \times B \times C_E$, where $B$ is the bank length for each category and $C_E$ is the channel number of feature embeddings.
Each category in the memory bank is updated with a ``First-In, First-Out'' strategy to ensure the bank contains the latest feature embeddings. The number of updated embeddings for each category at each iteration is set to $B_u$.
Then, to collect negative points, at each iteration, we evenly sample $\floor*{\frac{K_n}{|\mathcal{C}|}}$  points from each category's memory bank to form a total number of $K_n$ negative feature embeddings for contrastive learning.

Our proposed CBS strategy generates both category-balanced positive pairs and negative points.
It enables a more effective contrastive feature learning from the unlabeled point clouds, as shown later in Table~\ref{tab:exp_ab2}.

\subsection{Overall Architecture}

\vspace{-1.5mm}
\noindent 
{\bf Network architecture.} As shown in Fig.~\ref{fig:ssl_framework}, the overall framework of our semi-supervised method is composed of two branches, \ie, a supervised branch and an unsupervised branch. In each iteration, we sample labeled and unlabeled point clouds separately from the labeled and unlabeled set, and feed them into the network. For the backbone network, we apply the U-Net with sparse convolutions~\cite{3DSemanticSegmentationWithSubmanifoldSparseConvNet, choy20194d}, which is a top-performing network on several 3D segmentation datasets. In the supervised branch, the backbone network takes the labeled point cloud $\mathbf{P}^l \in \mathbb{R}^{N^l \times (3 + C_0)}$ with $3$D coordinates and $C_0$-dimensional  
%original 
raw features (\eg, colors) as input and generates point-wise features $\mathbf{F}^l \in \mathbb{R}^{N^l \times C_F}$, followed by a classifier to produce the semantic predictions $\mathbf{S}^l \in \mathbb{R}^{N^l \times |\mathcal{C}|}$, which are constrained by the cross-entropy loss as in Eq.~\eqref{eq:ce} with the ground-truth labels $\mathbf{Y}^l \in \mathbb{R}^{N^l}$.

For the unsupervised branch, we randomly augment the unlabeled point cloud $\mathbf{P}^u \in \mathbb{R}^{N^{u} \times (3 + C_0)}$ twice to produce a pair of training samples $\mathbf{P}^{u1} \in \mathbb{R}^{N^{u1} \times (3 + C_0)}$ and $\mathbf{P}^{u2} \in \mathbb{R}^{N^{u2} \times (3 + C_0)}$.
Then, we feed them into the backbone network to produce $\mathbf{F}^{u1}$ and $\mathbf{F}^{u2}$, and further employ a classifier to predict semantic scores $\mathbf{S}^{u1}$ and $\mathbf{S}^{u2}$, respectively.
Also, an additional projector is used to map $(\mathbf{F}^{u1}, \mathbf{F}^{u2})$ to feature embeddings $(\mathbf{E}^{u1}, \mathbf{E}^{u2})$ in the latent space. 
Without labels, the unsupervised branch is optimized by our proposed guided contrastive loss as in Eq.~\eqref{eq:main_loss}.

\vspace{1.5mm}
\noindent 
{\bf Overall objective.}
The overall objective of our semi-supervised framework is a combination of losses in the supervised and unsupervised branches: 
\vspace{-2mm}
\begin{equation}
L = L_l + \lambda L_u,
\end{equation}
\vspace{-5mm}

\noindent
where $L_u$ is our guided contrastive loss to enhance the feature learning with the unlabeled point clouds;
$L_l$ is a common cross-entropy loss for semantic segmentation; and $\lambda$ is a hyperparameter to adjust the loss ratio.

%%%%%%% experiments

\section{Experiments}

We present evaluations on our guided contrastive learning framework with both indoor and outdoor scenes. 
We use the mean Intersection-over-Union (mIoU) and mean accuracy (mAcc) as the evaluation metrics in the experiments.

\subsection{Experimental Setup}\label{sec:exp_setup}

\subsubsection{Datasets}\label{sec:dataset}
\vspace{-2mm}
We use both indoor (ScanNet V2~\cite{dai2017scannet} and S3DIS~\cite{armeni20163d}) and outdoor (SemanticKITTI~\cite{behley2019semantickitti}) datasets in our evaluations:  

\vspace*{0.5mm}
\noindent
{\bf ScanNet V2}~\cite{dai2017scannet} is a popular indoor 3D point cloud dataset that contains 1,613 3D scans with point-wise semantic labels. 
The whole data is split into a training set (1201 scans), a validation set (312 scans), and a testing set (100 scans). 
There are totally 20 categories for semantic segmentation. 

\vspace*{0.5mm}
\noindent
{\bf S3DIS}~\cite{armeni20163d} is another commonly-used indoor 3D point cloud dataset for semantic segmentation. It has 271 point cloud scenes across six areas, and there are in total 13 categories in the point-wise annotations. We follow the common split in previous works~\cite{NIPS2017_d8bf84be, li2018pointcnn} to utilize Area 5 as the validation set and adopt the other five areas as the training set.

\vspace*{0.5mm}
\noindent
{\bf SemanticKITTI}~\cite{behley2019semantickitti} is a large-scale outdoor point cloud dataset for 3D semantic segmentation in an autonomous driving scenario, where each scene is captured by the Velodyne-HDLE64 LiDAR sensor. The dataset contains 22 sequences that are divided into a training set (10 sequences with $\sim$19k frames), a validation set (1 sequence with $\sim$4k frames), and a testing set (11 sequences with $\sim$20k frames). There are 19 categories for semantic segmentation. 

\vspace*{0.5mm}
\noindent
{\bf SSL training set partition.}~
Following the conventional practice in SSL, we employ existing datasets in our evaluations and split the training set into labeled and unlabeled sets with five different ratios of labeled data,~\ie, \{5\%, 10\%, 20\%, 30\%, 40\%\}.
For SemanticKITTI, considering that adjacent frames could have very similar contents, when we split the dataset, we try our best to ensure that labeled and unlabeled data do not come from the same sequence. However, to achieve a specific labeled ratio, we may need to cut at most one sequence into two parts, the front part for labeled set and the latter part for unlabeled set. 

\vspace{-4mm}
\subsubsection{Augmentations for Semi-Supervised Learning}
\vspace{-2mm}
We adopt random crop as one of our augmentation operations. 
Since indoor and outdoor scenes have very different point distributions due to the use of different capturing devices, we perform different crop operations on them.

\vspace{0.5mm}
\noindent
{\bf Augmentations for indoor scenes.}~
For indoor scenes, the crop augmentation is implemented by randomly cropping a square region of size $3.5$m $\times$ $3.5$m in the top-down view.
For each unlabeled scene, we crop it twice and guarantee an overlap between the two cropped point clouds to build a point-to-point correspondence in the overlapping region.
Besides random crop, we adopt random rotation ($0$ - $2\pi$) around the z-axis (vertical axis) and random flip. 
Following the released code of~\cite{3DSemanticSegmentationWithSubmanifoldSparseConvNet}, we also adopt the elastic operation.

\vspace{0.5mm}
\noindent
{\bf Augmentations for outdoor scenes.}~
For outdoor scenes, we propose a sector-range crop centered at the origin that follows the beam pattern in LiDAR point clouds. Specifically, we randomize a heading angle in range $[0,2\pi]$ as the center direction of the sector and further randomize a field-of-view angle in range $[\frac{2}{3}\pi,2\pi]$ to form the cropping sector. For each unlabeled scene, two sectors are cropped with a guaranteed overlap for setting up a point-to-point correspondence.
Besides the sector-based crop, we adopt the commonly-used random flip, random rotation ($-\frac{\pi}{4}$  - $\frac{\pi}{4}$), and random scale (0.95 - 1.05) augmentations.

\vspace{-4mm}
\subsubsection{Implementation Details}\label{sec:imp}
\vspace{-2mm}

\noindent
{\bf Network details.}~
For both indoor and outdoor scenes, we utilize the sparse-convolution-based U-Net~\cite{3DSemanticSegmentationWithSubmanifoldSparseConvNet, choy20194d} as the backbone network for 3D semantic segmentation. The encoder applies sparse convolution layers with a stride of 2 to downsample the input volume six times, while the decoder gradually upsamples the volume back to the original size with six deconvolutions.  Submanifold sparse convolutions with a stride of 1 are used in the U-Net to encode the features. 
The projector is a multi-layer perception that maps the features to an embedding space.
For voxelizing the input point clouds, the voxel size is set to $2$cm for the indoor scenes and $10$cm for the outdoor scenes. 

\vspace{0.5mm}
\noindent
{\bf Training details.}~
For ScanNet V2, we train our SSL framework from scratch using an SGD optimizer.
The learning rate is initialized as 0.2 and decayed with the poly policy with a power of 0.9.  The batch size is 16, i.e., 16 labeled scenes and 16 unlabeled scenes.
For S3DIS, we apply the Adam optimizer with an initial learning rate of 0.02. 
We keep the same number of training iterations for different settings to train for $36$k iterations on ScanNet V2 and $8$k iterations on S3DIS using eight GPUs.
For a more stable semi-supervised training, we train the model with only supervised loss at the beginning 200 iterations. 
For outdoor scenes, the segmentation network is first pretrained on the labeled set by an Adam optimizer with a batch size of 48 and a learning rate of 0.02 for $16$k iterations on eight GPUs. Then, we train the network with our SSL framework on the labeled and unlabeled sets for another $18$k iterations by an Adam optimizer with a batch size of 24 and a learning rate of 0.002. The cosine annealing strategy is utilized to decay the learning rate. 
The loss ratio $\lambda$ for the guided contrastive loss is set to 0.1, while the temperature $\tau$ in the loss is set to 0.1. The confidence threshold $\gamma$ is 0.75.

\begin{table*}
	\begin{center}
		\scalebox{0.86}[0.83]{
			\begin{tabular}{c|c||cc|cc|cc|cc|cc||>{\columncolor{gray}}c>{\columncolor{gray}}c}
				\hline
				\multirow{2}{*}{Dataset} & \multirow{2}{*}{Model} & \multicolumn{2}{c|}{5\%} & \multicolumn{2}{c|}{10\%} & \multicolumn{2}{c|}{20\%} & \multicolumn{2}{c|}{30\%} & \multicolumn{2}{c||}{40\%} & \multicolumn{2}{c}{\cellcolor{gray}100\%}\\
				\hhline{~~|------------|}
				& & mIoU & mAcc & mIoU & mAcc & mIoU & mAcc & mIoU & mAcc & mIoU & mAcc & mIoU & mAcc\\
				\hline 
				\multirow{3}{*}{ScanNet V2} & Sup-only & 48.1 & 59.1 & 57.2 & 68.4 & 64.0 & 74.2 & 67.1 & 76.9 & 68.8 & 77.9 &72.9 &82.0\\
				& Semi-sup & 54.8 & 65.5 & 60.5 & 70.3 & 66.7 & 76.0 & 68.9 & 78.5 & 71.3 & 79.8 & 74.0 & 82.3 \\
				\hhline{~|-------------|}
				& \textbf{\textit{ Improv.}} & \textbf{+6.7} & \textbf{+6.4} & \textbf{+3.3} & \textbf{+1.9} & \textbf{+2.7} & \textbf{+1.8} & \textbf{+1.8} & \textbf{+1.6} & \textbf{+2.5} & \textbf{+1.9} & \textbf{+1.1} & \textbf{+0.3} \\
				\hline
				\multirow{3}{*}{S3DIS} & Sup-only & 45.0 & 57.9 & 52.9 & 62.7 & 59.9 & 67.9 & 61.2 & 69.2 & 62.6 & 69.4  & 66.4 & 73.1 \\
				& Semi-sup & 53.0 & 63.2 & 57.7 & 69.1 & 63.5 & 70.4 & 64.9 & 73.2 & 65.0 & 71.4 & 68.8 & 75.9\\
				\hhline{~|-------------|}
				& \textbf{\textit{ Improv.}} & \textbf{+8.0} & \textbf{+5.3} & \textbf{+4.8} & \textbf{+6.4} & \textbf{+3.6} & \textbf{+2.5} & \textbf{+3.7} & \textbf{+4.0} & \textbf{+2.4} & \textbf{+2.0} & \textbf{+2.4} & \textbf{+2.8} \\
				\hline
				\multirow{3}{*}{SemanticKITTI} & Sup-only & 34.8 & 40.0 & 43.9 & 55.2 & 53.8 & 62.1 & 55.4 & 63.6 & 57.4 & 65.6 & 65.0 & 72.1 \\
				& Semi-sup & 41.8 & 48.4 & 49.9 & 59.1 & 58.8 & 66.1 & 59.4 & 67.4 & 59.9 & 66.7 & 65.8 &	73.8 \\
				\hhline{~|-------------|}
				& \textbf{\textit{ Improv.}} & \textbf{+7.0} & \textbf{+8.4} & \textbf{+6.0} & \textbf{+3.9} & \textbf{+5.0} & \textbf{+4.0} & \textbf{+4.0} & \textbf{+3.8} & \textbf{+2.5} & \textbf{+1.1} & \textbf{+0.8} & \textbf{+1.7}\\
				\hline
			\end{tabular}
		}
	\end{center}
	\vspace{-2mm}
	\caption{Main results (mIoU(\%) and mAcc(\%)) on ScanNet V2~\cite{dai2017scannet} validation set, S3DIS~\cite{armeni20163d} Area 5 and SemanticKITTI~\cite{behley2019semantickitti} validation set with varying ratios \{5\%, 10\%, 20\%, 30\%, 40\%\} of labeled data. `Sup-only' means fully-supervised models trained with only labeled data, while `Semi-sup' represents our semi-supervised models. Particularly, in experiments with 100\% labeled data, the labeled set is also taken as input to the unsupervised branch in our `Semi-sup' models with our guided point contrastive loss as an auxiliary feature learning loss.}
	\vspace{-3mm}
	\label{tab:result1}
\end{table*}

\begin{figure}
	\vspace{-2mm}
	\centering
	\includegraphics[width=1.0\linewidth, height=5.2cm]{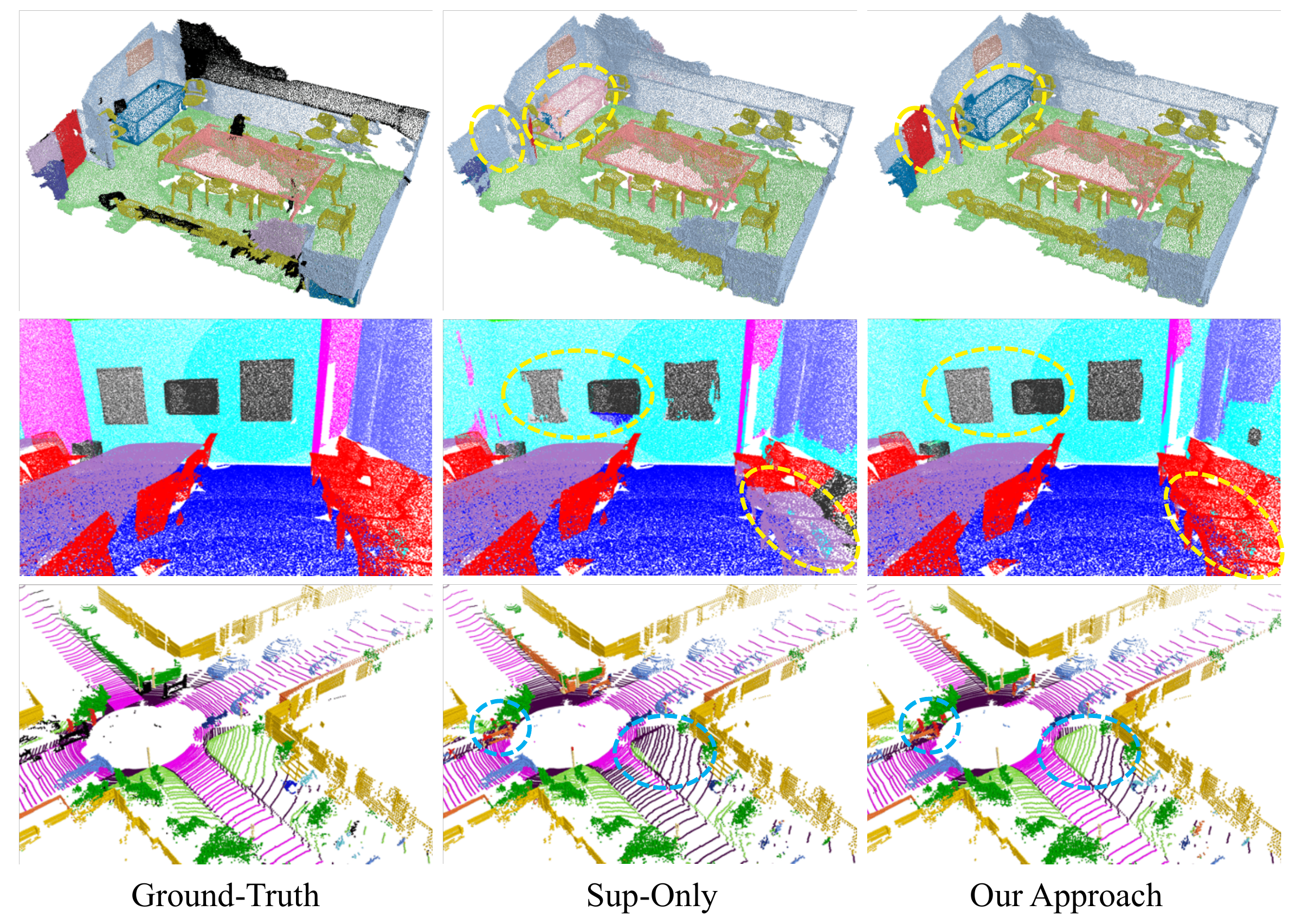}
	\vspace{-4mm}
	\caption{Qualitative results on indoor ScanNet V2 ($1^{st}$ row), indoor S3DIS ($2^{nd}$ row), and outdoor SemanticKITTI ($3^{rd}$ row). All models are trained with 20\% labeled data.
	}
	\vspace{-5mm}
	\label{fig:qualitative}
\end{figure}

\vspace{-1mm}
\subsection{Main Results}\label{sec:results}
\vspace{-1.5mm}
To demonstrate the effectiveness of our method on exploiting unlabeled data, 
we take the strong sparse-convolution-based  U-Net~\cite{3DSemanticSegmentationWithSubmanifoldSparseConvNet, choy20194d} as our backbone and follow the conventional practice in SSL to compare our semi-supervised models with models that are fully trained with only labeled point clouds, separately using \{5\%, 10\%, 20\%, 30\%, 40\%\} of the training set as the labeled data.

Table~\ref{tab:result1} summarizes the quantitative results on ScanNet V2, S3DIS, and SemanticKITTI in terms of mIoU and mAcc.
For all three datasets, both indoor and outdoor, our semi-supervised models consistently outperform the supervised-only ones for all ratios, showing 
that our SSL model is able to effectively leverage the unlabeled data to improve the embedding features and so the segmentation performance.
On all the datasets, the performance gap increases with the relative amount (ratio) of unlabeled data. 
Given 5\% labeled data and 95\% unlabeled data, our semi-supervised method improves the mIoU relatively by 13.9\%, 17.8\%, and 20.1\% in ScanNet V2, S3DIS, and SemanticKITTI, respectively. 
Further, we present some qualitative results in Fig.~\ref{fig:qualitative}, which shows 
that our model helps improve the segmentation quality with the unlabeled data.

Additionally, we conduct experiments on the 100\% ratio, in which the whole training set is taken as the labeled set and simultaneously fed also into the unsupervised branch as the unlabeled set. Our guided point contrastive loss serves as an auxiliary constraint for feature learning in the 100\% setting.
As shown in the last column of Table~\ref{tab:result1}, without extra unlabeled data, our method can still boost the network performance by enhancing the feature representation and model discriminative power via the guided contrastive learning in the unsupervised branch. 
We also compare our 100\% results with recent state-of-the-art methods in Table~\ref{tab:exp_sota}. Our supervised-only baseline model is already competitive among these methods, while our approach can further improve the prediction quality, which achieves excellent performance on all three datasets.

\noindent
\textbf{Transductive learning on 100\% ratio.}
Unlike inductive learning that aims at generalizing the model to unseen testing set, in transductive learning, the testing set is given and also observed in training. We extend the experiments on 100\% ratio to a transductive form by incorporating the testing data as part of the unlabeled data. We observe that the performance gets higher in the transductive form, as shown in Table~\ref{tab:trans}. The transductive model has comparable performance (70.2\%) as the SOTA’s in the SemanticKITTI CodaLab benchmark (Single Scan).

\begin{table}
	\small 
	\begin{center}
		\scalebox{0.9}{
			\setlength\tabcolsep{2.0pt}
			\begin{tabular}{c|c|cccccc}
				\hline
				\multirow{2}{*}{Method} & \multirow{2}{*}{Reference} & ScanNetV2 & S3DIS & \multicolumn{2}{c}{SemanticKITTI} \\
				& & Validation & Area 5 & Validation & Test \\
				\hline
				MinkowskiNet~\cite{choy20194d} & CVPR 2019 & 72.2	& 65.4 & 61.1 & 63.1\\
				KPConv~\cite{thomas2019kpconv} & ICCV 2019 & 69.2 & 67.1	& - & 58.8\\
				PointASNL~\cite{yan2020pointasnl} & CVPR 2020 & 66.4 & 62.6	& - & 46.8\\
				SPVNAS~\cite{tang2020searching}& ECCV 2020 & - & - & 64.7 & 66.4\\
				FusionNet~\cite{zhang2020} & ECCV 2020 & - & \textbf{67.2} & 63.7 & 61.3\\
				MVFusion~\cite{kundu2020virtual} & ECCV 2020 & \textbf{76.4} & 65.4 & - & -\\
				Cylinder3D~\cite{zhu2020cylindrical} & CVPR 2021	& -	& -	& \textbf{65.9} & \textbf{67.8}\\
				\hline
				Sup-only (100\%) & - & 72.9 & 66.4 & 65.0 & 65.4\\
				Our model (100\%)& - & \textbf{74.0} & \textbf{68.8} & \textbf{65.8} & \textbf{67.7}\\
				\hline
			\end{tabular}
		}
	\end{center}
	\vspace{-2mm}
	\caption{Semantic segmentation results (mIoU(\%)) on ScanNet validation set, S3DIS Area 5, and SemanticKITTI validation and test set. The top two results are highlighted for all datasets.
		We re-implement the strong sparse-convolution-based U-Net~\cite{3DSemanticSegmentationWithSubmanifoldSparseConvNet} as our baseline `Sup-only' model. Our model further uses our guided point contrastive loss as an auxiliary feature learning loss. All our results are based on the 100\% labeled ratio.}
	\vspace{-5mm}
	\label{tab:exp_sota}
\end{table}

\vspace{-1mm}
\subsection{Ablation Studies}\label{sec:ab_study}
\vspace{-1mm}

\noindent
\textbf{Pseudo guidance \& CBS.}~
We conduct ablation studies on our pseudo guidance and CBS design using the 20\% labeled data on ScanNet V2.
Table~\ref{tab:exp_ab1} shows the contribution of each component in our method. Without the pseudo guidance, the effect of point contrastive loss in exploiting unlabeled data is limited. 
By avoiding potential intra-category negative pairs,
our pseudo-label guidance provides the most significant performance gain (1.5\%) over the vanilla point contrastive loss. Our confidence guidance, which promotes the feature learning quality, further improves the mIoU from 65.9\% to 66.4\%. Our full model with CBS to enhance feature diversity leads to the highest performance 66.7\%. 

\vspace{0.5mm}
\noindent
\textbf{CBS on different datasets.}
For further analyzing the effect of our CBS, we compare CBS with random sampling on ScanNet V2 (indoor) and SemanticKITTI (outdoor) with 20\% labeled data.
Table~\ref{tab:exp_ab2} reports the results. 
Compared with indoor scenes, outdoor data suffers more from the category imbalance problem.
We count the number of points in each category of the training set for both datasets.
In SemanticKITTI, 6 out of the 19 categories, \ie, `bicycle', `motorcycle', `person', `bicyclist', `motorcyclist', and `traffic-sign', have fewer than 1\textperthousand~points 
in the training set. The most rare `motorcyclist' category has only 0.04\textperthousand. However, in ScanNet V2 training set, the point category distribution is more balanced; the most rare category `sink' still has 2.75\textperthousand~points.  Hence, our CBS contributes a larger increase for SemanticKITTI. For a category with very sparse points, it is hard to be sampled with random sampling.  
CBS increases the probability of selecting samples from these categories and thus improves the feature diversity. 

\begin{table}
	\renewcommand\arraystretch{1.0}
	\begin{center}
		\scalebox{0.82}{
			\begin{tabular}{c||c|c|c}
				\hline
				\multirow{2}{*}{Dataset} &  \multirow{2}{*}{ \tabincell{c}{Sup-only\\Baseline}} & \multicolumn{2}{c}{Semi-sup} \\
				\cline{3-4}
				& & \tabincell{c}{Inductive} & \tabincell{c}{Transductive} \\
				\hline
				ScanNet V2 Val. & 72.9 & 74.0 & \textbf{74.7}\\
				S3DIS Area 5 & 66.4 & 68.8 & \textbf{69.7} \\
				SemanticKITTI Val. & 65.0 & 65.8 & \textbf{66.5} \\
				SemanticKITTI Test & 65.4 & 67.7 & \textbf{70.2} \\
				
				\hline
			\end{tabular}
		}
	\end{center}
	\vspace{-2mm}
	\caption{Extensive experiments on 100\% ratio for transductive learning, 
		where the testing set is incorporated into the unlabeled set for training. The evaluation metric is mIoU(\%).}
	\vspace{-3mm}
	\label{tab:trans}
\end{table}

\begin{table}
	\small 
	\begin{center}
		\scalebox{0.92}[0.92]{
			\begin{tabular}{c|ccc|c}
				\hline
				\tabincell{c}{Setting}& \tabincell{c}{Pseudo-label\\Guidance} & \tabincell{c}{Confidence\\Guidance} & CBS &  mIoU(\%) \\
				\hline
				Sup-only &&&& 64.0 \\
				\hline 
				PointInfoNCE &&&& 64.4\\
				Our model-a & \checkmark & & & 65.9  \\
				Our model-b & \checkmark & \checkmark &  & 66.4  \\
				Our full model & \checkmark & \checkmark & \checkmark & 66.7  \\
				\hline
			\end{tabular}
		}
	\end{center}
	\vspace{-2mm}
	\caption{Effects of different components in our SSL approach for 3D scene semantic segmentation, where `CBS' denotes our category-balanced sampling strategy.
		All experiments are conducted on ScanNet V2 with 20\% labeled data.}
	\label{tab:exp_ab1}
	\vspace{-4mm}
\end{table}

\vspace{0.5mm}
\noindent
\textbf{Partial views vs. random crop.}
PointContrast~\cite{xie2020pointcontrast} suggests that the multi-view design is critical in improving the quality of the pretrained model. For a scene in ScanNet V2, the multi-view design samples two partial views of the scene instead of cropping the reconstructed point cloud. We also try this multi-view strategy in our unsupervised branch. However, the experimental results on ScanNet V2 (with 20\% labeled data) show that applying partial views instead of random crop even lowers the performance from 66.7\% to 63.2\%. The possible reason for the performance drop is that the multi-view design could widen the discrepancy between the labeled and unlabeled sets, introducing imprecise semantic predictions in the unsupervised branch. 

\vspace{0.5mm}
\noindent
\textbf{The projector.}
The projector is essential for contrastive learning, as discussed in~\cite{chen2020simple}. If we remove the projector, 
the performance (mIoU) drops from 66.7\% to 65.0\% on \text{ScanNet} V2 dataset with 20\% labeled data.

\begin{table}
	\small 
	\begin{center}
		\scalebox{0.99}[0.95]{
			\begin{tabular}{c|ccc}
				\hline
				Sampling Strategy & ScanNet V2 &  SemanticKITTI \\
				\hline
				Random & 66.4 & 57.1 \\
				CBS & 66.7 & 58.8 \\
				\hline
			\end{tabular}
		}
	\end{center}
	\vspace{-2mm}
	\caption{Effects of category-balanced sampling on ScanNet V2 and SemanticKITTI with 20\% labeled data (metric: mIoU(\%)).}
	\vspace{-2mm}
	\label{tab:exp_ab2}
\end{table}

\begin{table}
	\small 
	\begin{center}
		\scalebox{0.92}{
			\setlength\tabcolsep{3.0pt}
			\begin{tabular}{c|c|ccccc}
				\hline
				Strategy & \tabincell{c}{Sup-\\Only} & MSE & \tabincell{c}{Cosine\\Similarity} & \tabincell{c}{PointInfoNCE} & \tabincell{c}{Self-\\Training} & Ours \\
				\hline
				mIoU(\%) & 64.0 & 65.0 & 64.9 & 64.4 & 65.5 & 66.7 \\
				\hline
			\end{tabular}
		}
	\end{center}
	\vspace{-2mm}
	\caption{Comparing different strategies for semi-supervised 3D semantic segmentation on ScanNet V2 with 20\% labeled data.}
	\label{tab:exp_ab3}
	\vspace{-5mm}
\end{table}

\vspace{-1mm}
\subsection{Analysis on various Semi-supervised Strategies}\label{sec:exp_strategy}
\vspace{-1.5mm}
Apart from our guided contrastive learning, we also experiment with some other semi-supervised strategies, and the performance comparison is shown in Table~\ref{tab:exp_ab3}.

\vspace{1mm}
\noindent \textbf{Consistency regularization.} With consistency regularization as the semi-supervised strategy, we apply the MSE loss as in Eq.~\eqref{eq:mse} or the cosine similarity loss to align the matched features under different perturbations.  The results surpass the supervised-only model by 1.0\% and 0.9\%, respectively. Our method achieves better performance with an improvement of 2.7\% with stronger constraints on the features by not only keeping the consistency between matched points but also pushing away the points with different semantic predictions in the embedding space. 

\vspace{1mm}
\noindent \textbf{PointInfoNCE loss.} Directly applying the PointInfoNCE loss in PointContrast~\cite{xie2020pointcontrast} as the unsupervised loss does not perform well in the semi-supervised setting. The high probability of sampling negative pairs within the same category 
adversely affects the feature learning.

\vspace{1mm}
\noindent \textbf{Self-training.} 
Pseudo-label-based self-training is another alternative strategy for SSL. 
We have also tried it by first training a model with the labeled set, loading it to produce pseudo labels for the unlabeled set, and then constraining the semantic predictions in the unsupervised branch with the pseudo labels by a cross-entropy loss. Self-training also performs well in exploiting unlabeled data (65.5\%), but our method can still attain a higher performance (66.7\%).

%%%%%%% conclusion
\vspace{-1mm}
\section{Conclusion}
\vspace{-1mm}
In this work, we present a semi-supervised framework to take advantage of unlabeled point clouds to perform 3D semantic segmentation in a data-efficient manner.
With our guided point contrastive loss, the network can learn more discriminative features by leveraging our pseudo-label and confidence guidance. 
Also, we propose the category-balanced sampling to benefit contrastive learning with more diverse feature embeddings.
Experimental results show the effectiveness of our approach to exploit unlabeled 3D data and to improve the model generalization ability.

\vspace{1.2mm}
\noindent \textbf{Acknowledgments} The project is supported in part by the Research Grants Council of the Hong Kong Special Administrative Region (Project no. CUHK 14206320).

%------------------------------------------------------------------------

{\small
\bibliographystyle{ieee_fullname}
\bibliography{egbib}

\begin{thebibliography}{10}\itemsep=-1pt

\bibitem{armeni20163d}
Iro Armeni, Ozan Sener, Amir~R. Zamir, Helen Jiang, Ioannis Brilakis, Martin
  Fischer, and Silvio Savarese.
\newblock {3D} semantic parsing of large-scale indoor spaces.
\newblock In {\em CVPR}, 2016.

\bibitem{behley2019semantickitti}
Jens Behley, Martin Garbade, Andres Milioto, Jan Quenzel, Sven Behnke, Cyrill
  Stachniss, and Jurgen Gall.
\newblock Semantickitti: A dataset for semantic scene understanding of lidar
  sequences.
\newblock In {\em ICCV}, 2019.

\bibitem{chen2020simple}
Ting Chen, Simon Kornblith, Mohammad Norouzi, and Geoffrey Hinton.
\newblock A simple framework for contrastive learning of visual
  representations.
\newblock In {\em ICML}, 2020.

\bibitem{NEURIPS2020_fcbc95cc}
Ting Chen, Simon Kornblith, Kevin Swersky, Mohammad Norouzi, and Geoffrey~E
  Hinton.
\newblock Big self-supervised models are strong semi-supervised learners.
\newblock In {\em NeurIPS}, 2020.

\bibitem{choy20194d}
Christopher Choy, JunYoung Gwak, and Silvio Savarese.
\newblock {4D} spatio-temporal convnets: Minkowski convolutional neural
  networks.
\newblock In {\em CVPR}, 2019.

\bibitem{dai2017scannet}
Angela Dai, Angel~X. Chang, Manolis Savva, Maciej Halber, Thomas Funkhouser,
  and Matthias Nie{\ss}ner.
\newblock Scan{N}et: Richly-annotated {3D} reconstructions of indoor scenes.
\newblock In {\em CVPR}, 2017.

\bibitem{french2020semi}
Geoffrey French, Samuli Laine, Timo Aila, Michal Mackiewicz, and Graham
  Finlayson.
\newblock Semi-supervised semantic segmentation needs strong, varied
  perturbations.
\newblock In {\em BMVC}, 2020.

\bibitem{3DSemanticSegmentationWithSubmanifoldSparseConvNet}
Benjamin Graham, Martin Engelcke, and Laurens van~der Maaten.
\newblock {3D} semantic segmentation with submanifold sparse convolutional
  networks.
\newblock In {\em CVPR}, 2018.

\bibitem{NIPS2004_96f2b50b}
Yves Grandvalet and Yoshua Bengio.
\newblock Semi-supervised learning by entropy minimization.
\newblock In {\em NeurIPS}, 2005.

\bibitem{hadsell2006dimensionality}
Raia Hadsell, Sumit Chopra, and Yann LeCun.
\newblock Dimensionality reduction by learning an invariant mapping.
\newblock In {\em CVPR}, 2006.

\bibitem{he2020momentum}
Kaiming He, Haoqi Fan, Yuxin Wu, Saining Xie, and Ross Girshick.
\newblock Momentum contrast for unsupervised visual representation learning.
\newblock In {\em CVPR}, 2020.

\bibitem{hjelm2018learning}
R.~Devon Hjelm, Alex Fedorov, Samuel Lavoie-Marchildon, Karan Grewal, Phil
  Bachman, Adam Trischler, and Yoshua Bengio.
\newblock Learning deep representations by mutual information estimation and
  maximization.
\newblock In {\em ICLR}, 2019.

\bibitem{hou2020exploring}
Ji Hou, Benjamin Graham, Matthias Nie{\ss}ner, and Saining Xie.
\newblock Exploring data-efficient {3D} scene understanding with contrastive
  scene contexts.
\newblock In {\em CVPR}, 2021.

\bibitem{hung2018adversarial}
Wei-Chih Hung, Yi-Hsuan Tsai, Yan-Ting Liou, Yen-Yu Lin, and Ming-Hsuan Yang.
\newblock Adversarial learning for semi-supervised semantic segmentation.
\newblock In {\em BMVC}, 2018.

\bibitem{jiang2019hierarchical}
Li Jiang, Hengshuang Zhao, Shu Liu, Xiaoyong Shen, Chi-Wing Fu, and Jiaya Jia.
\newblock Hierarchical point-edge interaction network for point cloud semantic
  segmentation.
\newblock In {\em ICCV}, 2019.

\bibitem{jiang2020pointgroup}
Li Jiang, Hengshuang Zhao, Shaoshuai Shi, Shu Liu, Chi-Wing Fu, and Jiaya Jia.
\newblock Point{G}roup: Dual-set point grouping for {3D} instance segmentation.
\newblock In {\em CVPR}, 2020.

\bibitem{ke2020guided}
Zhanghan Ke, Kaican~Li Di~Qiu, Qiong Yan, and Rynson~W.H. Lau.
\newblock Guided collaborative training for pixel-wise semi-supervised
  learning.
\newblock In {\em ECCV}, 2020.

\bibitem{NEURIPS2020_d89a66c7}
Prannay Khosla, Piotr Teterwak, Chen Wang, Aaron Sarna, Yonglong Tian, Phillip
  Isola, Aaron Maschinot, Ce Liu, and Dilip Krishnan.
\newblock Supervised contrastive learning.
\newblock In {\em NeurIPS}, 2020.

\bibitem{kundu2020virtual}
Abhijit Kundu, Xiaoqi Yin, Alireza Fathi, David Ross, Brian Brewington, Thomas
  Funkhouser, and Caroline Pantofaru.
\newblock Virtual multi-view fusion for 3d semantic segmentation.
\newblock In {\em ECCV}, 2020.

\bibitem{lahoud20193d}
Jean Lahoud, Bernard Ghanem, Marc Pollefeys, and Martin~R Oswald.
\newblock {3D} instance segmentation via multi-task metric learning.
\newblock In {\em ICCV}, 2019.

\bibitem{temporalICLR2017}
Samuli Laine and Timo Aila.
\newblock Temporal ensembling for semi-supervised learning.
\newblock In {\em ICLR}, 2017.

\bibitem{landrieu2018large}
Loic Landrieu and Martin Simonovsky.
\newblock Large-scale point cloud semantic segmentation with superpoint graphs.
\newblock In {\em CVPR}, 2018.

\bibitem{lee2013pseudo}
Dong-Hyun Lee.
\newblock Pseudo-label: The simple and efficient semi-supervised learning
  method for deep neural networks.
\newblock In {\em Workshop on challenges in representation learning, ICML},
  2013.

\bibitem{li2018pointcnn}
Yangyan Li, Rui Bu, Mingchao Sun, Wei Wu, Xinhan Di, and Baoquan Chen.
\newblock {PointCNN}: Convolution on $\chi$-transformed points.
\newblock In {\em NeurIPS}, 2018.

\bibitem{maturana2015voxnet}
Daniel Maturana and Sebastian Scherer.
\newblock Vox{N}et: A {3D} convolutional neural network for real-time object
  recognition.
\newblock In {\em IROS}, 2015.

\bibitem{mittal2019semi}
Sudhanshu Mittal, Maxim Tatarchenko, and Thomas Brox.
\newblock Semi-supervised semantic segmentation with high-and low-level
  consistency.
\newblock {\em TPAMI}, 2019.

\bibitem{miyato2018virtual}
Takeru Miyato, Shin-ichi Maeda, Masanori Koyama, and Shin Ishii.
\newblock Virtual adversarial training: a regularization method for supervised
  and semi-supervised learning.
\newblock {\em TPAMI}, 2018.

\bibitem{oord2018representation}
Aaron van~den Oord, Yazhe Li, and Oriol Vinyals.
\newblock Representation learning with contrastive predictive coding.
\newblock {\em arXiv preprint arXiv:1807.03748}, 2018.

\bibitem{ouali2020semi}
Yassine Ouali, C{\'e}line Hudelot, and Myriam Tami.
\newblock Semi-supervised semantic segmentation with cross-consistency
  training.
\newblock In {\em CVPR}, 2020.

\bibitem{qi2017pointnet}
Charles~R. Qi, Hao Su, Kaichun Mo, and Leonidas~J. Guibas.
\newblock Point{N}et: Deep learning on point sets for {3D} classification and
  segmentation.
\newblock In {\em CVPR}, 2017.

\bibitem{NIPS2017_d8bf84be}
Charles~R. Qi, Li Yi, Hao Su, and Leonidas~J. Guibas.
\newblock {PointNet++}: Deep hierarchical feature learning on point sets in a
  metric space.
\newblock In {\em NeurIPS}, 2017.

\bibitem{NIPS2015_378a063b}
Antti Rasmus, Mathias Berglund, Mikko Honkala, Harri Valpola, and Tapani Raiko.
\newblock Semi-supervised learning with ladder networks.
\newblock In {\em NeurIPS}, 2015.

\bibitem{riegler2017octnet}
Gernot Riegler, Ali Osman~Ulusoy, and Andreas Geiger.
\newblock Oct{N}et: Learning deep {3D} representations at high resolutions.
\newblock In {\em CVPR}, 2017.

\bibitem{shi2019pointrcnn}
Shaoshuai Shi, Xiaogang Wang, and Hongsheng Li.
\newblock {PointRCNN}: {3D} object proposal generation and detection from point
  cloud.
\newblock In {\em CVPR}, 2019.

\bibitem{simonovsky2017dynamic}
Martin Simonovsky and Nikos Komodakis.
\newblock Dynamic edge-conditioned filters in convolutional neural networks on
  graphs.
\newblock In {\em CVPR}, 2017.

\bibitem{song2017semantic}
Shuran Song, Fisher Yu, Andy Zeng, Angel~X. Chang, Manolis Savva, and Thomas
  Funkhouser.
\newblock Semantic scene completion from a single depth image.
\newblock In {\em CVPR}, 2017.

\bibitem{souly2017semi}
Nasim Souly, Concetto Spampinato, and Mubarak Shah.
\newblock Semi supervised semantic segmentation using generative adversarial
  network.
\newblock In {\em ICCV}, 2017.

\bibitem{tang2020searching}
Haotian Tang, Zhijian Liu, Shengyu Zhao, Yujun Lin, Ji Lin, Hanrui Wang, and
  Song Han.
\newblock Searching efficient 3d architectures with sparse point-voxel
  convolution.
\newblock In {\em ECCV}, 2020.

\bibitem{NIPS2017_68053af2}
Antti Tarvainen and Harri Valpola.
\newblock Mean teachers are better role models: Weight-averaged consistency
  targets improve semi-supervised deep learning results.
\newblock In {\em NeurIPS}, 2017.

\bibitem{thomas2019kpconv}
Hugues Thomas, Charles~R. Qi, Jean-Emmanuel Deschaud, Beatriz Marcotegui,
  Fran{\c{c}}ois Goulette, and Leonidas~J. Guibas.
\newblock {KPConv}: Flexible and deformable convolution for point clouds.
\newblock In {\em ICCV}, 2019.

\bibitem{wang20203dioumatch}
He Wang, Yezhen Cong, Or Litany, Yue Gao, and Leonidas~J. Guibas.
\newblock {3DIoUMatch}: Leveraging iou prediction for semi-supervised {3D}
  object detection.
\newblock In {\em CVPR}, 2021.

\bibitem{wang2019dynamic}
Yue Wang, Yongbin Sun, Ziwei Liu, Sanjay~E. Sarma, Michael~M. Bronstein, and
  Justin~M. Solomon.
\newblock Dynamic graph {CNN} for learning on point clouds.
\newblock {\em Acm Transactions On Graphics (TOG)}, 2019.

\bibitem{wu2019pointconv}
Wenxuan Wu, Zhongang Qi, and Li Fuxin.
\newblock Point{C}onv: Deep convolutional networks on {3D} point clouds.
\newblock In {\em CVPR}, 2019.

\bibitem{xie2020pointcontrast}
Saining Xie, Jiatao Gu, Demi Guo, Charles~R. Qi, Leonidas~J. Guibas, and Or
  Litany.
\newblock Point{c}ontrast: Unsupervised pre-training for {3D} point cloud
  understanding.
\newblock In {\em ECCV}, 2020.

\bibitem{yan2020sparse}
Xu Yan, Jiantao Gao, Jie Li, Ruimao Zhang, Zhen Li, Rui Huang, and Shuguang
  Cui.
\newblock Sparse single sweep lidar point cloud segmentation via learning
  contextual shape priors from scene completion.
\newblock In {\em AAAI}, 2021.

\bibitem{yan2020pointasnl}
Xu Yan, Chaoda Zheng, Zhen Li, Sheng Wang, and Shuguang Cui.
\newblock Pointasnl: Robust point clouds processing using nonlocal neural
  networks with adaptive sampling.
\newblock In {\em CVPR}, 2020.

\bibitem{ye2019unsupervised}
Mang Ye, Xu Zhang, Pong~C. Yuen, and Shih-Fu Chang.
\newblock Unsupervised embedding learning via invariant and spreading instance
  feature.
\newblock In {\em CVPR}, 2019.

\bibitem{zhai2019s4l}
Xiaohua Zhai, Avital Oliver, Alexander Kolesnikov, and Lucas Beyer.
\newblock {S4L}: Self-supervised semi-supervised learning.
\newblock In {\em ICCV}, 2019.

\bibitem{zhang2020}
Feihu Zhang, Jin Fang, Benjamin Wah, and Philip Torr.
\newblock Deep fusionnet for point cloud semantic segmentation.
\newblock In {\em ECCV}, 2020.

\bibitem{zhao2019pointweb}
Hengshuang Zhao, Li Jiang, Chi-Wing Fu, and Jiaya Jia.
\newblock Point{W}eb: Enhancing local neighborhood features for point cloud
  processing.
\newblock In {\em CVPR}, 2019.

\bibitem{zhao2020sess}
Na Zhao, Tat-Seng Chua, and Gim~Hee Lee.
\newblock {SESS}: Self-ensembling semi-supervised {3D} object detection.
\newblock In {\em CVPR}, 2020.

\bibitem{zhu2020cylindrical}
Xinge Zhu, Hui Zhou, Tai Wang, Fangzhou Hong, Yuexin Ma, Wei Li, Hongsheng Li,
  and Dahua Lin.
\newblock Cylindrical and asymmetrical 3d convolution networks for lidar
  segmentation.
\newblock In {\em CVPR}, 2021.

\bibitem{zou2021pseudoseg}
Yuliang Zou, Zizhao Zhang, Han Zhang, Chun-Liang Li, Xiao Bian, Jia-Bin Huang,
  and Tomas Pfister.
\newblock Pseudo{S}eg: Designing pseudo labels for semantic segmentation.
\newblock In {\em ICLR}, 2021.

\end{thebibliography}
}

\end{document}